\documentclass[letterpaper, 10 pt, conference]{ieeeconf}  

\IEEEoverridecommandlockouts                              

\usepackage{graphicx}
\usepackage{booktabs} 
\usepackage{amsmath}
\usepackage{amsfonts}
\usepackage{bm} 
\usepackage{tabularx}
\usepackage[normalem]{ulem}

\usepackage{xcolor}
\usepackage{hyperref}

\title{Establishing Appropriate Trust via Critical States}

\author{Sandy H. Huang, Kush Bhatia, Pieter Abbeel, Anca D. Dragan \\
University of California, Berkeley, EECS}

\begin{document}

\newcommand{\eref}[1]{Eqn. (\ref{#1})}
\newcommand{\sref}[1]{Sec. \ref{#1}}
\newcommand{\figref}[1]{Fig. \ref{#1}}
\newcommand{\tabref}[1]{Table \ref{#1}}

\newcommand\Tstrut{\rule{0pt}{2.6ex}}         
\newcommand\Bstrut{\rule[-0.9ex]{0pt}{0pt}}   

\newcommand{\adnote}[1]{
 {\textcolor{blue}{\textbf{Anca: #1}}}}
\newcommand{\shnote}[1]{
 {\textcolor{red}{\textbf{SH: #1}}}}

\newcommand{\prg}[1]{\noindent\textbf{#1. }} 

\maketitle

\begin{abstract}
In order to effectively interact with or supervise a robot, humans need to have an accurate mental model of its capabilities and how it acts. Learned neural network policies make that particularly challenging. We propose an approach for helping end-users build a mental model of such policies. Our key observation is that for most tasks, the essence of the policy is captured in a few \emph{critical states}: states in which it is very important to take a certain action.  Our user studies show that if the robot shows a human what its understanding of the task's critical states is, then the human can make a more informed decision about whether to deploy the policy, and if she does deploy it, when she needs to take control from it at execution time.
\end{abstract}
\section{Introduction}
When humans have an accurate mental model of a robot, their subsequent interactions with this robot are safer and more seamless. This mental model may include the robot's intentions~\cite{Gielniak_2011,Dragan_2013,Szafir_2014}, its objectives~\cite{Huang_2017}, its capabilities~\cite{Nikolaidis_2017,Kwon_2018}, or its decision-making process~\cite{Wang_2016}.

In particular, giving human end-users an accurate mental model of a robot's capabilities is key to establishing an appropriate level of trust in the robot~\cite{Dzindolet_2003,Lee_2004,Ososky_2013}. Establishing \emph{appropriate} levels of trust in robots is essential: if end-users do not trust a robot, they may unnecessarily interfere with its operation, and will fail to take advantage of all its capabilities~\cite{Freedy_2007}. On the other hand, if end-users over-trust a robot, they will expect it to act correctly in situations that it in fact cannot handle, which leads to unexpected behavior, and perhaps injuries and damage. As robots become more capable, they may unintentionally lead humans to over-trust them~\cite{Cha_2015}. In general, trust is a complex phenomenon, and there are a variety of ways in which robots and machines may influence human end-users' trust~\cite{Muir_1987,McKnight_2001,Lewis_2018}. 

Establishing appropriate trust is particularly challenging when the robot has learned a complex black-box policy. For instance, recently neural network policies have been trained to perform robotic manipulation skills~\cite{Levine_2016} and drive in the real world~\cite{Bojarski_2016}. These neural networks are trained end-to-end to map directly from raw inputs (e.g., images) to a distribution over actions to take. 
To decide how much to trust a learned policy, we have to know whether the robot has figured out the correct actions to take. But, it is impossible to examine what the robot plans to do in every possible state.

\begin{figure}[t!]
    \centering
    \includegraphics[width=0.8\linewidth]{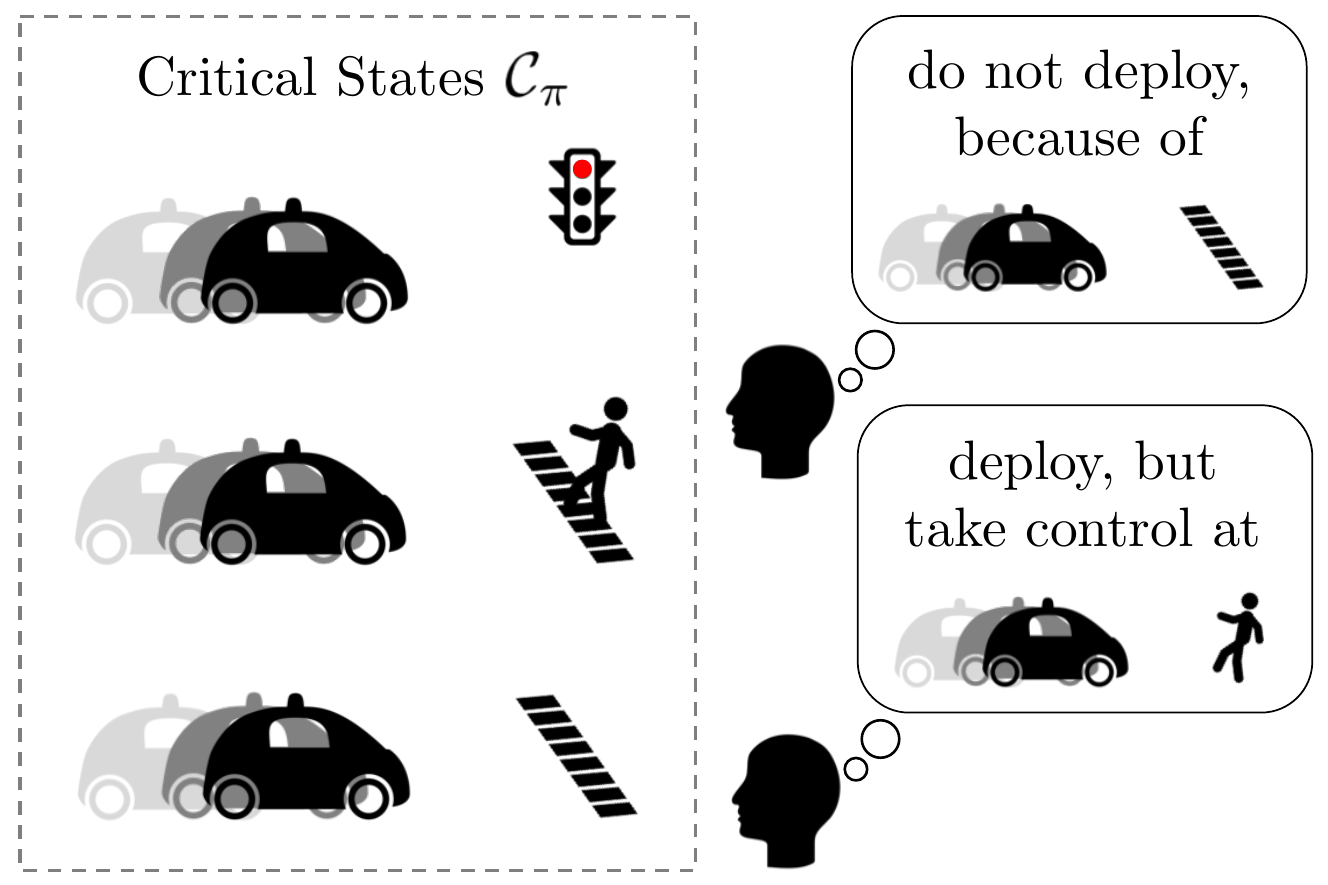}
    \caption{By introducing human end-users to a policy $\pi$'s critical states $\mathcal{C}_\pi$ (left), we enable them to make a more informed decision about whether to deploy the policy, and when to take control from it. For example, suppose that a self-driving car's policy believes it is critical to stop when it encounters red lights, a pedestrian crossing a crosswalk, and an empty crosswalk. An end-user (top right) might see these critical states and decide not to ride in this car, because the last critical state is clearly incorrect. A different end-user (bottom right) might be comfortable riding in this car, but will be more aware of possibly needing to take control when there is a pedestrian crossing the road without a crosswalk in sight, because the policy did not consider that to be a critical state.}
    \label{fig:frontfig}
    \vspace{-1em}
\end{figure}

Our insight is that the end-user does not need to know what the robot would do in \emph{all} states. For many tasks, in most states the ultimate outcome of the task is similar, regardless of which action the robot takes locally. But there are a few states---\emph{critical states}---where it really matters which action the robot takes. 

For instance, imagine an autonomous car driving down a highway. When there are no vehicles nearby, it does not matter whether the car maintains its current speed, speeds up or slows down slightly, or turns slightly to the right or left. In contrast, if the vehicle directly in front slams on its brakes, the autonomous car must immediately slow down as well. The latter is a critical state, whereas the former is not.

Usually when end-users are introduced to a robot, they are only told summary statistics of this robot's performance. For instance, a potential passenger may be told that a particular autonomous car has driven more than a million miles, without causing any accidents. Without more information, this passenger has no way of knowing what kinds of states this car still cannot handle. If she expects the autonomous car to do the right thing in a critical state, and it does not, then it may be too late to recover.

As end-users observe and interact with a robot over time, they will gradually improve their mental model of it~\cite{Dragan_2014}, just as they do when observing other humans~\cite{Baker_2009,Jara_2016}. However, it may take a while for end-users to learn a sufficiently-accurate mental model in this way. The hope is that we can speed up this process by exposing humans to more informative examples of the robot's behavior.

To this end, we propose showing end-users how the robot acts in critical states, to give them a better understanding of what it has learned, and enable them to decide which situations to trust the robot in (\figref{fig:frontfig}). After seeing how a robot acts in critical states, a potential user may decide that this robot is not trustworthy, and \emph{decline to use it}. Or, in human-in-the-loop setups---for instance, a passenger riding in a self-driving car, or an engineer supervising robot arms in a factory---this ensures users are well-equipped to decide \emph{when they need to take control} over the robot's operation.

Our main contribution is a method for \emph{algorithmic assurance}~\cite{Israelsen_2017}, that enables end-users to more quickly establish an appropriate level of trust in robots that they interact with, rely on, or supervise. Our user studies suggest that humans are indeed able to develop more appropriate trust in a robot through observing how it acts in what it considers to be critical states, compared to just observing it act over time. We evaluate this through both self-reported measures of trust, as well as through allowing users to take control during execution of the policy~\cite{Freedy_2007}: if they have developed an appropriate level of trust, they would only choose to take control in critical states that the robot likely cannot handle.
\section{Background}
\subsection{Notation}
We consider the setting of a Markov Decision Process (MDP), defined by $\{\mathcal{S}, \mathcal{A}, \mathcal{P}, \mathcal{R}, \gamma\}$, where $\mathcal{S}$ is the state space, $\mathcal{A}$ the action space, $\mathcal{P}: \mathcal{S} \times \mathcal{A} \times \mathcal{S} \to \mathbb{R}$ the transition probabilities, $\mathcal{R}: \mathcal{S} \times \mathcal{A} \times \mathcal{S} \to \mathbb{R}$ the reward function, and $\gamma \in (0,1]$ the discount factor.

A robot's policy $\pi$ is a stochastic function mapping each state to a distribution over actions ($\pi: \mathcal{S} \to \Delta_{\mathcal{A}}$, where $\Delta_{\mathcal{A}}$ is the probability simplex on $\mathcal{A}$). Its value function at state $s$ is
\begin{equation}
    V^\pi(s) = \max_a \int_{s'} P(s,a,s') [R(s,a,s') + \gamma V^\pi(s')],
\end{equation}
and its action-value function at state $s$ and taking action $a$ is
\begin{equation}
    Q^\pi(s,a) = \int_{s'} P(s,a,s') [R(s,a,s') + \gamma \max_{a'} Q^\pi(s',a')].
\end{equation}

In this framework, a critical state $s$ is one for which $Q^\pi(s,a)$ varies greatly across different actions $a$: there are a small number of actions for which $Q^\pi(s,\cdot)$ is high, but for most actions it is mediocre or low. We will define this formally in the next section.

\subsection{Maximum-Entropy Reinforcement Learning}
Typically a robot's goal is to maximize expected cumulative discounted reward, or return:
\begin{equation}
    \mathbb{E}_{\pi,\mathcal{P}} \left[ \sum_t \gamma^t R(s_t,a_t,s_{t+1}) \right].
\end{equation}
Depending on the MDP, this may result in policies that are essentially deterministic, treating all states as critical.

In contrast, in maximum-entropy reinforcement learning, the policy is trained to not only maximize return, but also to act as randomly as possible while doing so~\cite{Ziebart_2008,Haarnoja_2017a,Haarnoja_2017}. Concretely, the policy is trained to maximize
\begin{equation}
    \mathbb{E}_{\pi,\mathcal{P}} \left[ \sum_t \gamma^t [R(s_t,a_t,s_{t+1}) + \alpha \mathcal{H}(\pi(\cdot|s_t))] \right],
\end{equation}
where $\alpha$ determines the tradeoff between maximizing return and entropy, and $\mathcal{H}(\pi(\cdot|s_t))$ is the entropy of the policy's output action distribution at state $s_t$. 
This leads to a policy with meaningful critical states, since it learns to acts randomly in states where the action has little impact on return, and to act purposefully in states where the action does have a major impact on return.

We train our neural network policies using Soft Actor-Critic\footnote{We use the implementation at \url{github.com/haarnoja/sac}.} (SAC)~\cite{Haarnoja_2017}, a deep reinforcement learning method that is based on maximum entropy reinforcement learning. We find that in practice, training with SAC indeed produces policies with meaningful critical states.
\begin{figure*}[h!]
    \centering
    \includegraphics[width=0.95\linewidth]{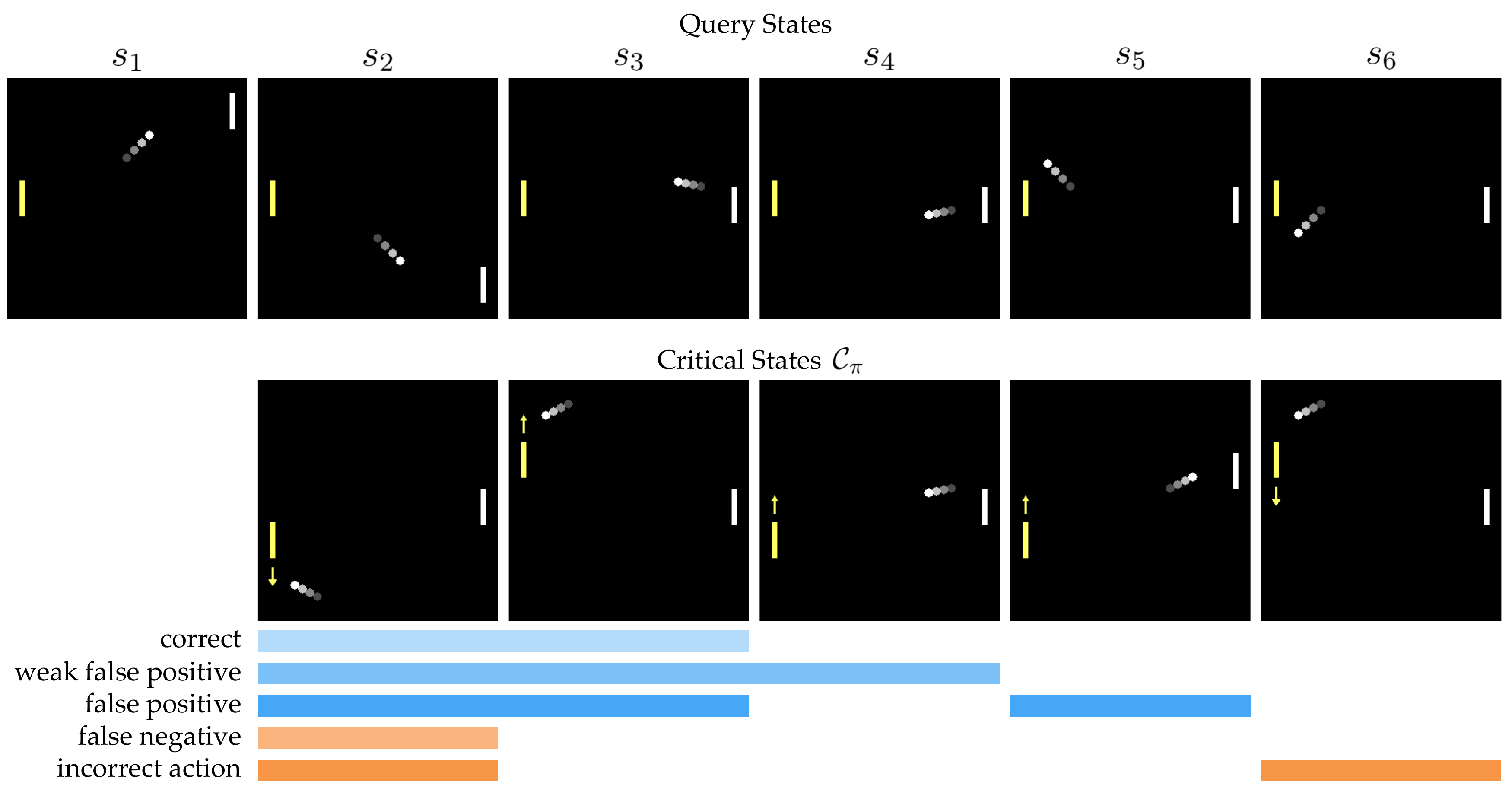}
    \caption{The query states and sets of critical states $\mathcal{C}_\pi$ shown in our user study for Pong. The policy controls the yellow paddle. Query states $s_1$ through $s_4$ are not critical, because the paddle has plenty of time to reach the ball, whereas $s_5$ and $s_6$ are. The colored bars indicate which states are included in each possible $\mathcal{C}_\pi$. For example, the \emph{incorrect-action} $\mathcal{C}_\pi$ contains one correct critical state (the leftmost one) and one incorrect-action critical state (the rightmost one). The \emph{false-negative} $\mathcal{C}_\pi$ contains one correct critical state, but is missing the second correct critical state---so the corresponding policy would likely miss balls heading toward it from above. Each possible $\mathcal{C}_\pi$ contains at least one correct critical state (the leftmost one).}
    \label{fig:pongstates}
    \vspace{-1em}
\end{figure*}

\section{Computing and Using Critical States}\label{sec:criticalstates}

\subsection{Computation of Critical States}
\label{subsec:computecpi}

\prg{Policy-Based} Recall that critical states are those in which a policy (or human) greatly prefers a small set of possible actions over all others. A natural definition of the set of critical states $\mathcal{C}_\pi$ for a stochastic policy $\pi$ is thus
\begin{equation}
    \mathcal{C}_\pi = \{s \, | \, \mathcal{H}(\pi(\cdot|s)) < t \},
\end{equation}
where $\mathcal{H}(\pi(\cdot|s))$ is the entropy of the policy's output action distribution at state $s$, and $t\in\mathbb{R}$ is the threshold for being considered ``critical.''
This definition of critical states can be applied to both continuous and discrete action spaces.

\prg{Value-Based} Certain reinforcement learning approaches for training policies, such as actor-critic methods, also learn a value or action-value function in parallel to (or instead of) learning a policy~\cite{Sutton_1998}. Value functions capture the long-term consequences of a policy's actions, so when they are available, they are a reasonable alternative for computing critical states.

If we define critical states more concretely as those in which acting randomly will produce a much worse result than acting optimally, then the set of critical states $\mathcal{C}_\pi$ for a stochastic policy $\pi$ is:
\begin{equation}
    \mathcal{C}_\pi = \{s \, | \, (\max_a Q^{\pi}(s,a) - \frac{1}{|\mathcal{A}|} \sum_{a} Q^{\pi}(s,a)) > t \},
\end{equation}
where $Q^\pi$ is the learned action-value function. If the action space is continuous, this can be applied after discretization. Computing critical states based on a learned value function $V^\pi$ is also possible, by using one-step rollouts to estimate $Q^\pi$ for each action. 

We train our policies with SAC, which learns a policy and an action-value function in parallel. In practice, we found that computing critical states based on action-value functions was more reliable, because the policy may learn to exploit environment characteristics (e.g., action clipping) to maximize entropy.

Note that with either of these two approaches, computing the critical states of a policy is agnostic to the implementation of the policy itself; only access to either the policy's or action-value function's output is required, so this can be directly applied to black-box policies.

\subsection{Using Critical States}

We assume a human expert at the task. Let $\mathcal{C}_h$ be the set of (ground-truth) states that she considers critical. We do not know what $\mathcal{C}_h$ is---and, in fact, this may differ across human end-users---so we cannot check whether $\mathcal{C}_\pi$ and $\mathcal{C}_h$ are the same. However, what we can do is expose the human to $\mathcal{C}_\pi$. Below, we describe the interaction we envision.

\noindent\textbf{Decline to deploy due to false positives, false negatives, or incorrect actions. }Before using a robot that has learned a policy $\pi$, the human end-user first gets to observe its actions in the states \emph{it} considers as critical, $\mathcal{C}_\pi$. If the human spots false-positive or false-negative critical states (i.e., states that are in $\mathcal{C}_\pi$ but not in $\mathcal{C}_h$ or vice versa), then she can decline to deploy the robot. False-negative critical states happen, for instance, when an autonomous car does not realize that stopping for a red light is a critical state. False-positive critical states happen, for instance, when an autonomous car considers it critical to slow down, even if there is quite a bit of space left to the car in front.
Both false-negative and false-positive critical states indicate that the robot has failed to learn something fundamental about the task, and thus perhaps should not be trusted. Similarly, if the policy identifies a true-positive critical state but is mistaken about which action is correct in that state, then the end-user will observe that and not trust the policy as a result.

\noindent\textbf{Take control.} We are also interested in the case where $\mathcal{C}_\pi$ does not have any \emph{obvious} false-positive, false-negative, or incorrect-action critical states, and the user decides to go ahead and deploy the robot, but the robot operates with the user in the loop. At execution time the user is able to take control from the policy whenever she deems it necessary. Because she has already observed how the policy acts for states in $\mathcal{C}_\pi$, the user is better equipped to take control from the policy when necessary, and refrain from doing so when not necessary.

\subsection{Justification of Critical States}
The user should have enough information based on critical states to take control when necessary at execution time.
Note that at execution time, any state $s$ encountered by the robot must fall into one of three cases: (1) $s \not\in \mathcal{C}_h$, (2) $s \in \mathcal{C}_h$ and $s \in \mathcal{C}_\pi$, or (3) $s \in \mathcal{C}_h$ and $s \not\in \mathcal{C}_\pi$.

In case (1), the user does not consider this state to be critical, so by definition she does not care which action the policy chooses and will refrain from taking control. In contrast, in cases (2) and (3), the user does consider this state to be critical, and cares about which action the policy takes. Since the user has observed (and approved) the policy's actions for states in $\mathcal{C}_\pi$, she should trust the robot in case (2). In case (3), $s$ is a false-negative critical state that the end-user forgot about when approving this policy. Since this is a critical state that the policy does not know is critical, she should take control from the policy immediately.

If the user had not been able to observe how the policy acts for states in $\mathcal{C}_\pi$, then she would not be able to distinguish between when she absolutely must take control (states in case (3)), and when she should not but may be tempted to (states in case (2)).
\begin{figure*}[ht!]
    \centering
    \includegraphics[width=0.95\linewidth]{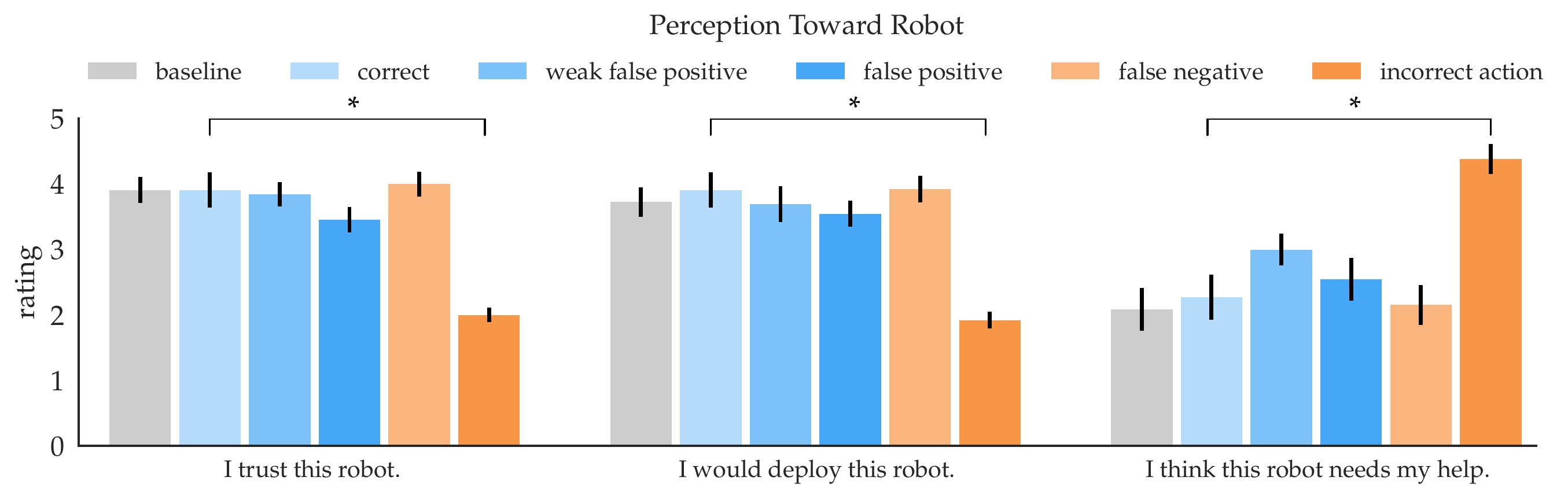}
    \caption{Ratings for Likert statements in \sref{sec:pongexp}, averaged across participants in each condition. Higher ratings mean higher agreement.}
    \label{fig:pong_subj}
\end{figure*}

\begin{figure*}[ht!]
    \centering
    \includegraphics[width=0.98\linewidth]{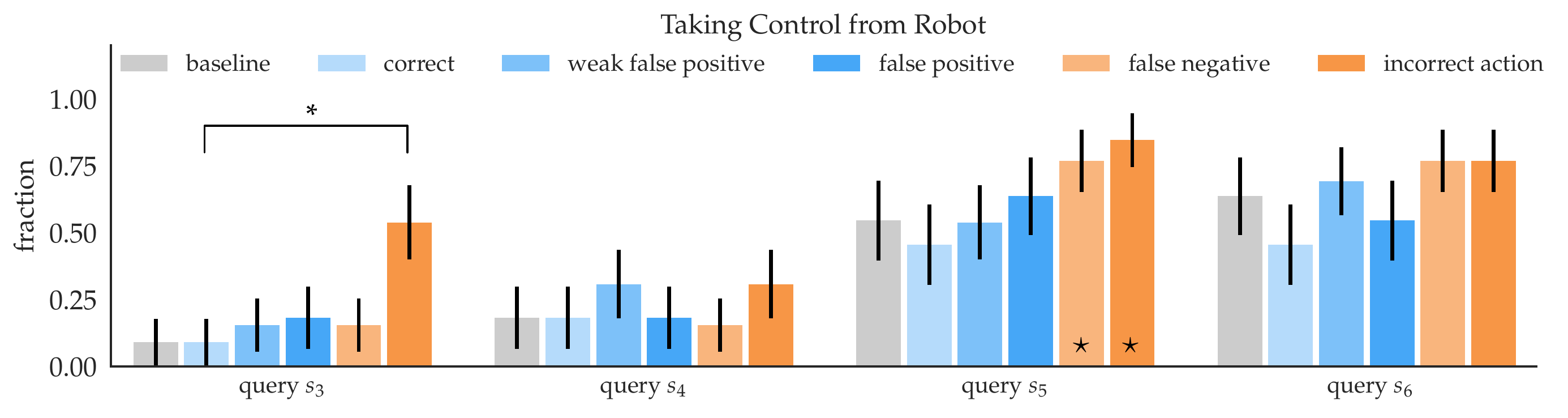}
    \caption{Participants' yes/no responses for whether they would take control of the policy at a particular query state (from \figref{fig:pongstates}). A $\star$ indicates that this is a state in which participants should choose to take control, based on the critical states they observed. Results for $s_1$ and $s_2$ are omitted---people overwhelmingly chose to not take control, regardless of which condition they were in.}
    \label{fig:pong_obj}
    \vspace{-1em}
\end{figure*}

\section{User Study: Impact of Critical States}
\label{sec:pongexp}

We begin by investigating how human end-users draw conclusions after observing the critical states of a policy, and how they respond to different errors (i.e., false positives, false negatives, or incorrect actions) in these critical states. In order to explore this in a systematic way, instead of obtaining critical states from trained policies, we construct sets of critical states where each set has at most one error. From this we can learn, for example, how much seeing a false-positive critical state impacts trust, versus seeing an incorrect-action. Later, in our main user study (\sref{sec:drivingexp}), we expose end-users to critical states from actual trained policies.

\subsection{Experiment Design}
The study consists of three phases. In the \emph{query} phase, we first introduce participants to the task and ask them, for a handful of states, whether they consider it critical to take a particular action in that state (\figref{fig:pongstates}, top row). This is to get a sense of what $\mathcal{C}_h$ is across participants. In the \emph{exposure} phase, we introduce participants to a policy, for instance by showing them its critical states. Finally, in the \emph{test} phase, we ask participants whether they would take control from the policy, for each of the same states as in the \emph{query} phase.

\prg{Domain} We chose a straightforward task with clear critical states: Pong. In Pong, a ball bounces back and forth between two paddles, and the goal is to use your paddle to hit the ball past your opponent's. So states in which the ball is headed back toward your opponent are non-critical, since it does not matter much how you move your paddle. In contrast, states in which the ball is heading toward your paddle and has almost reached it are critical.

\prg{Manipulated Variables} We manipulate the set of critical states $\mathcal{C}_\pi$ shown to the participant. We construct five options for $\mathcal{C}_\pi$---\emph{correct}, \emph{false-negative}, \emph{weak-false-positive}, \emph{false-positive}, and \emph{incorrect-action}---that cover all the possible problems with a particular policy's critical states (\figref{fig:pongstates}). In the baseline condition, instead of showing the participant a set of critical states, we simply give them a summary statistic of the robot's performance: ``this policy wins in 95\% of cases.'' This establishes a baseline of how much participants trust policies for Pong that are reasonably good.

\prg{Dependent Measures} We are interested in whether observing a set of critical states leads participants to develop \emph{appropriate} trust in the policy that generated those critical states. We measure trust in two ways: subjectively with five-point Likert questions, and objectively with which \emph{test} phase states participants choose to take control from the policy in, and whether those are correct (i.e., in $C_h$ and either not in $C_\pi$, or in $C_\pi$ but as an incorrect action). This \emph{test} phase simulates execution-time: after the end-user has already chosen to deploy the policy, and is now supervising it.

\prg{Hypothesis}

\prg{H1} When $\mathcal{C}_\pi$ contains false-negative, false-positive, or incorrect-action critical states, users are less inclined to trust the policy $\pi$, compared to if its critical states match $\mathcal{C}_h$ perfectly (i.e., the \emph{correct} condition).

\prg{H2} In states that are critical (i.e., in $\mathcal{C}_h$), participants will take control if a policy $\pi$'s critical states $\mathcal{C}_\pi$ suggest that this policy will not choose the correct action in this state.  For example, since the \emph{false-negative} $\mathcal{C}_\pi$ for Pong is missing critical states in which the paddle needs to immediately move upward to hit the ball, this should lead participants to take control in similar states at execution time (e.g., query state $s_5$). But, they should not take control at state $s_6$, since the \emph{false-negative} $\mathcal{C}_\pi$ includes a similar critical state and chooses the right action.

\prg{Subject Allocation} We used a between-subjects design. We ran this experiment on a total of 72 participants across the six conditions, recruited via Amazon Mechanical Turk. The average age of the participants was 31.4 ($SD=6.7$). The gender ratio was 0.32 female.

\subsection{Analysis}

\prg{Subjective} We asked participants how much they trust the robot, whether they would deploy it, and whether they thought the robot needed their help (\figref{fig:pong_subj}).

We found a significant difference between \emph{incorrect-action} and \emph{correct} for all three subjective measures (Student's t test, $p < 0.0001$). However, \emph{false-positives} and \emph{false-negatives} did not decrease users' perception compared to \emph{correct} (the trend is in the right direction for the false positives). This may be because Pong is a relatively simple domain, which makes humans more inclined to give policies the benefit of the doubt, in terms of being able to generalize to other critical states (in the case of the \emph{false-negative} $\mathcal{C}_\pi$).

\prg{Objective} We also asked participants, for each of the six query states (\figref{fig:pongstates}), a yes/no question for whether they would take control of the policy at that state (\figref{fig:pong_obj}). In the \emph{query} phase, participants agreed that of the six states, only $s_5$ and $s_6$ are truly critical (i.e., in $\mathcal{C}_h$). We see that overall, across all conditions, participants tend to take control in these two critical states, and not in the others. This supports our assumption that humans will tend to only take control of policies in states that are within $\mathcal{C}_h$.

However, this also indicates that participants are taking control even when it is not necessary. For instance, users who saw the \emph{correct} $\mathcal{C}_\pi$ saw it act correctly in states similar to both critical query states, but still almost half of users choose to take control in that state. 

On the bright side, we saw a number of trends in line with our hypothesis. First, we do notice that for these two critical query states, users tend to be less likely to take control after seeing \emph{correct} $\mathcal{C}_\pi$, compared to just being told a summary statistic about the policy, in the baseline condition.

Second, participants in the \emph{incorrect-action} condition again indicated low trust in the robot, by choosing to take control more often, even in state $s_3$, which is only weakly critical.  We found participants chose to take control significantly more in the \emph{incorrect-action} condition than the \emph{correct} condition for $s_3$ (Student's t, $p < 0.01$) and $s_5$ ($p = 0.05$).

Third, participants who saw false-positive and false-negative critical states actually tended to take control more often than those who saw correct ones, suggesting that they did pick up somewhat on the problems indicated by $\mathcal{C}_\pi$ (with weak significance, for $s_5$ and $s_6$, $p = 0.11$).

\noindent\textbf{Summary.} Overall, participants responded most strongly to critical states that reveal incorrect actions. There, they would intervene before deployment. For false negatives, they would tend to take control away from the robot more compared to participants who saw correct critical states. False positives only benefited from slight improvements in how much participants would take control, though at the same time false positives are the smallest of errors, as we discussed in \sref{sec:criticalstates}.

\begin{figure*}[ht!]
    \centering
    \includegraphics[width=0.95\linewidth]{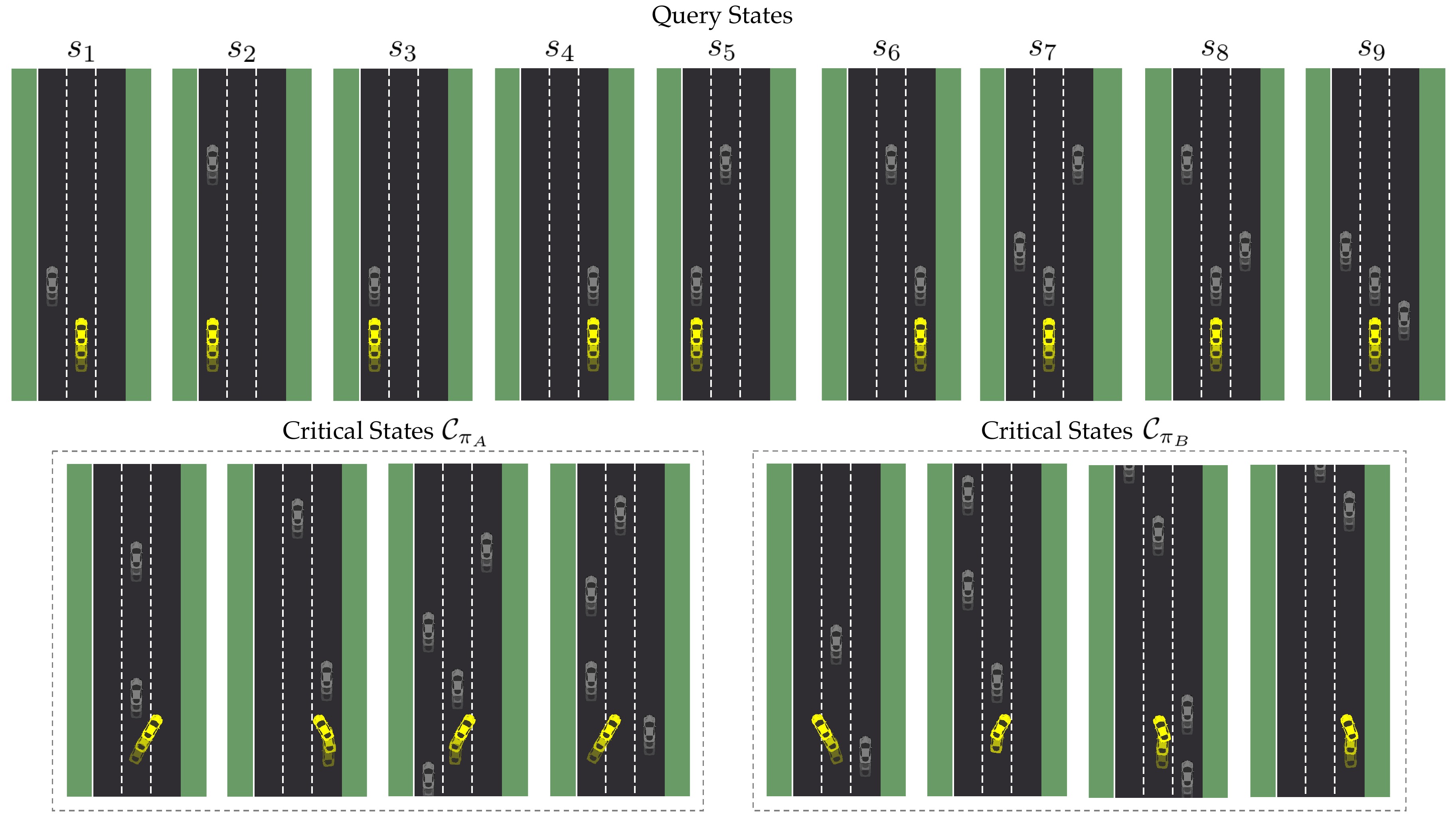}
    \caption{The query states and a subset of the ten critical states $\mathcal{C}_\pi$ shown in our main user study. The policy controls the steering of the yellow car. Query states $s_1$ and $s_2$ are not critical, but the rest are.}
    \label{fig:drivingstates}
    \vspace{-1em}
\end{figure*}

\begin{figure*}[ht!]
    \centering
    \includegraphics[width=0.95\linewidth]{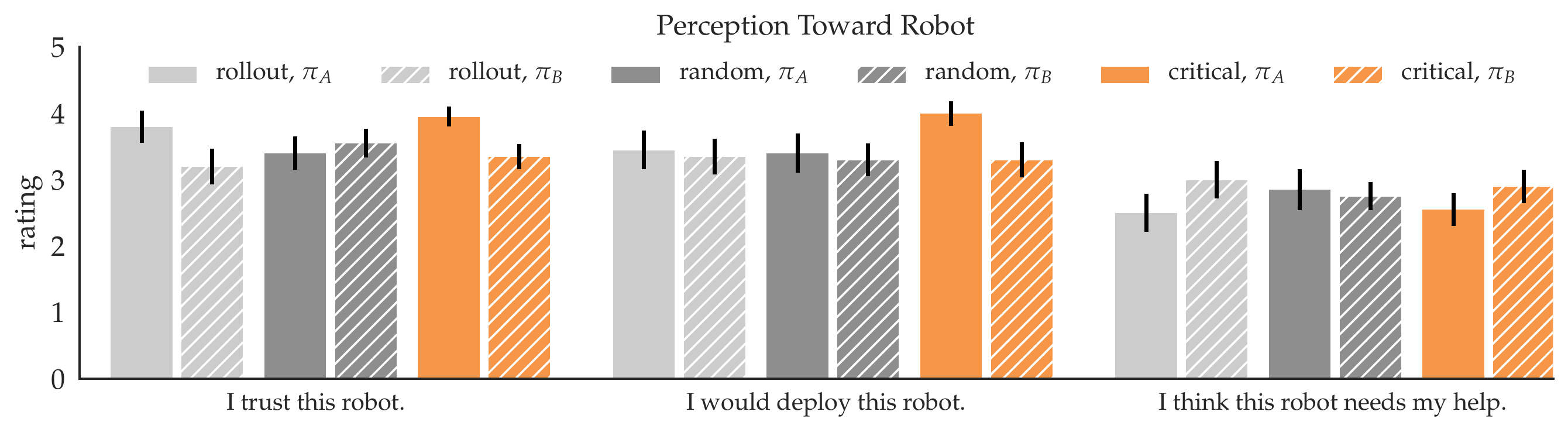}
    \caption{Ratings for Likert statements in \sref{sec:drivingexp}, averaged across participants in each condition. Higher ratings mean higher agreement.}
    \label{fig:driving_subj}
    \vspace{-1em}
\end{figure*}
\begin{figure}[ht!]
    \centering
    \includegraphics[width=0.95\linewidth]{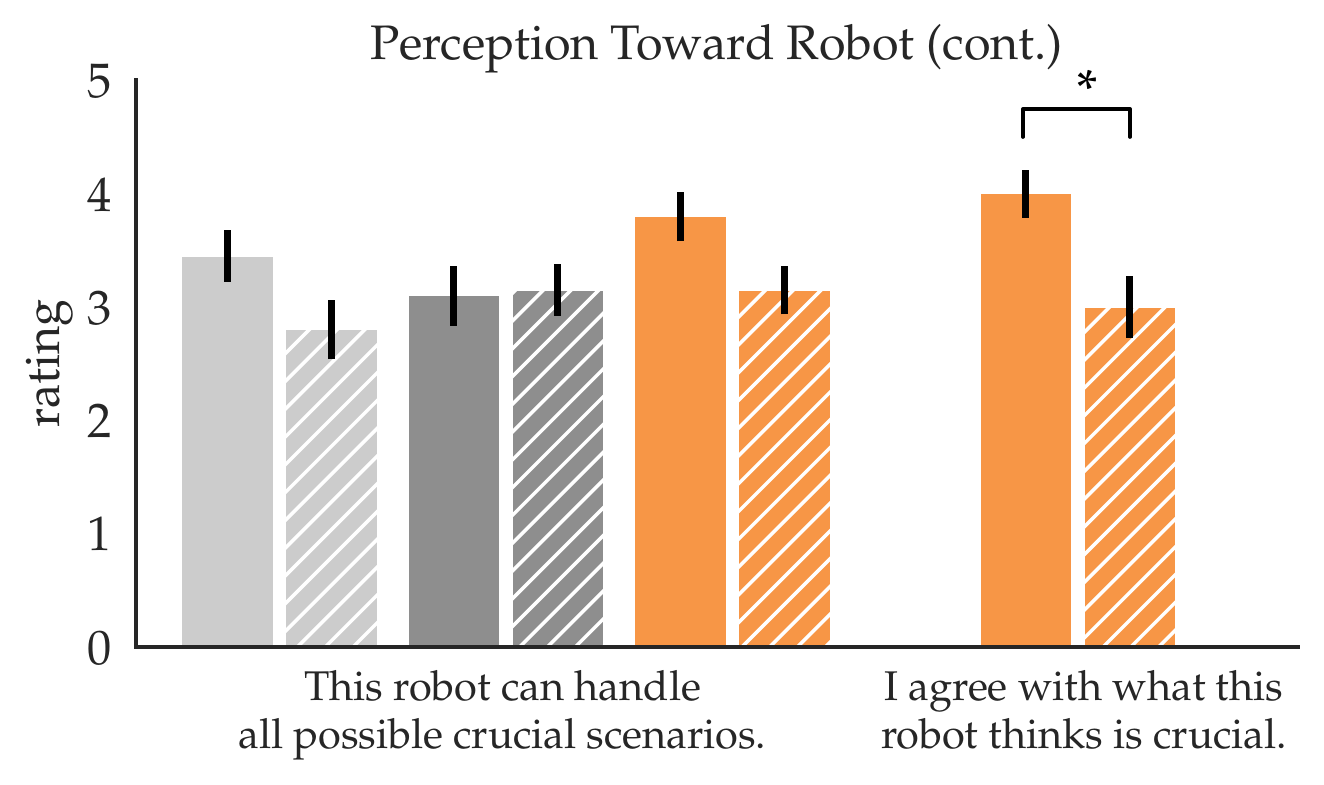}
    \caption{Ratings for Likert statements in \sref{sec:drivingexp}.}
    \label{fig:driving_subj2}
    \vspace{-1em}
\end{figure}
\begin{figure*}[ht!]
    \centering
    \includegraphics[width=0.95\linewidth]{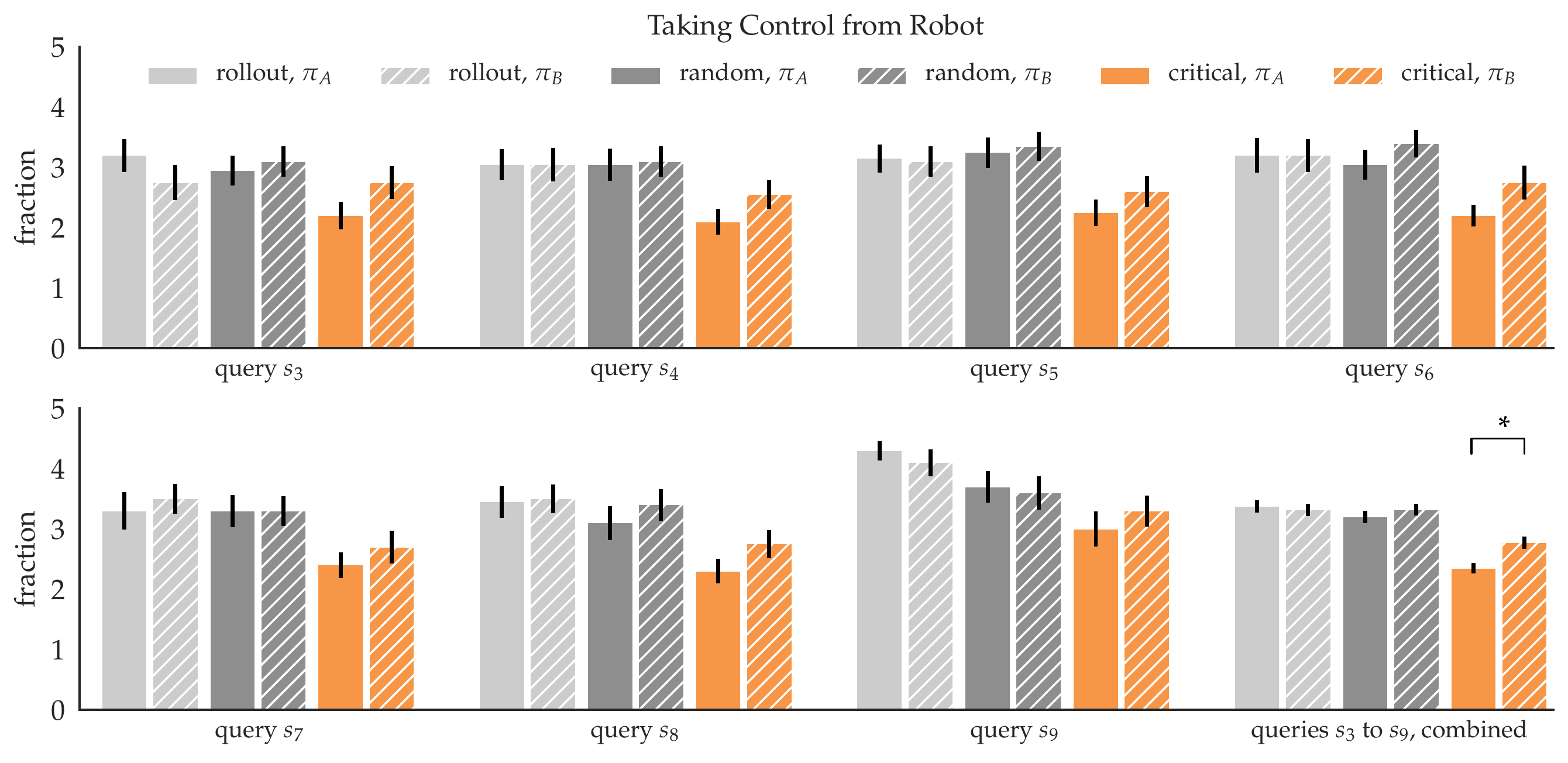}
    \caption{Participants' responses for whether they would take control of the policy at a particular query state (from \figref{fig:drivingstates}). Results for $s_1$ and $s_2$ are omitted, since people overwhelmingly chose not to take control, regardless of which condition they were in.}
    \label{fig:driving_obj}
    \vspace{-1em}
\end{figure*}

\section{User Study: Utility of Critical States}
\label{sec:drivingexp}

Our previous study analyzed how people respond to different errors that critical states might reveal. In our main user study, we evaluate the utility of showing the critical states of a policy $\pi$ against other options of exposing end-users to the policy, in terms of establishing appropriate trust. 

We train two neural network policies for a driving domain, and hypothesize that critical states are best at helping people figure out which one is better. We train these policies using SAC, and use Gaussian mixture policies with four components~\cite{Haarnoja_2017}.

In practice, critical states in $\mathcal{C}_\pi$ may be very similar to each other, so instead of showing all states in $\mathcal{C}_\pi$ to the human, we first cluster these states (with k-means++) and then show the policy's behavior in the most critical state from each cluster. We take advantage of the fact that neural network policies learn hidden-layer feature representations, and use the output of the last hidden layer as features for clustering. Concretely, we collect 10,000 timesteps by rolling out each policy, cluster the 10\% most critical states into ten clusters, and show the most critical state from each cluster. So, we end up showing ten critical states per policy.

\subsection{Experimental Design}
This study consists of the same three phases as the previous study.

\prg{Domain} We train policies to drive in a top-down driving simulator that mimics highway driving. The goal of the policy is to navigate down this road while passing other, slower cars. Car dynamics follow the bicycle vehicle model~\cite{Taheri_1990}. The state space consists of an indicator for which lane the robot car is currently in, its position and heading, and the relative positions, heading, and speed of other nearby cars. The action space is continuous and one-dimensional, in the range $[-1,1]$; it corresponds to the change in steering angle.\footnote{We discretize this action space evenly into 200 possible actions, in order to compute critical states using the learned action-value function.} The reward function encourages forward progress and penalizes getting close to other cars, being off-center in the lane, turning, and steering sharply.

\prg{Manipulated Variables} We manipulate two variables: how a user is exposed to the policy, and the quality of the policy. For exposure type, we compare our approach of showing critical states to two baselines: showing a one-minute rollout of the policy, and showing how the policy acts in random states, rather than critical ones. These two baselines are meant to approximate the states a user would happen to encounter as she observes and interacts with the robot over time.

For the quality of the policy, we have a policy $\pi_A$ trained for 10,000 iterations, and another policy $\pi_B$ that is trained for only 3,000 iterations. Both policies achieve similar performance on the task: $\pi_A$ averages one crash per 700 timesteps, and $\pi_B$ averages one crash per 640 timesteps.\footnote{Note that since the agent can only steer, and the other cars surrounding the agent are all driving slower than it, it will often encounter situations where crashes are inevitable.} But $\pi_B$ fails in a few simple traffic scenarios, that $\pi_A$ has learned to navigate successfully---including query states $s_5$ and $s_7$ (\figref{fig:drivingstates}).

\figref{fig:drivingstates} shows a subset of the ten critical states per policy. Looking closely at the critical states of policy $\pi_B$ (\figref{fig:drivingstates}), we see the rightmost two states are false-positives, whereas all the critical states of policy $\pi_A$ look reasonable. On average, the critical states of policy $\pi_B$ are also of simpler driving scenarios, which suggests that it may not be able to handle more challenging ones.

\prg{Dependent Measures} We keep the same dependent measures as in the previous user study (\sref{sec:pongexp}), except we add two Likert questions that ask participants more specifically about how much they trust the policy with respect to critical states, and change the yes/no question for taking control to a five-point Likert question where higher means more likely to take control.

\prg{Hypothesis} Showing users the critical states of a policy establishes appropriate trust, compared to other approaches of exposing users to policies. Appropriate trust, in this setting, means that participants trust $\pi_A$ over $\pi_B$, both in their Likert responses and in how often they choose to take control from the policy.

\prg{Subject Allocation} We used a between-subjects design for exposure type, and within-subjects for policy quality to reduce variance. We ran this experiment on a total of 60 participants across the three conditions, recruited via Amazon Mechanical Turk. The average age of the participants was 32.5 ($SD=6.7$). The gender ratio was 0.27 female.

\subsection{Analysis}
\prg{Subjective}  We see that across all five questions, users who have seen the \emph{critical states} of both policies tend to favor policy $\pi_A$, the better one (\figref{fig:driving_subj}, \figref{fig:driving_subj2}). This trend is also visible for participants who see a one-minute rollout of each policy, but not as consistently. In contrast, when participants see how the policy acts in randomly-selected states, they rate policies $\pi_A$ and $\pi_B$ similarly, indicating that their trust is incorrectly calibrated.

We ran a two-way repeated-measures ANOVA, with exposure and policy quality as factors and user ID as a random effect, for each item except the question on agreement. We observe a weak interaction effect between exposure and policy quality for the question on trust (F(2,57) = 2.37, $p = 0.1$). We also ran a post-hoc Tukey HSD for each item, which confirmed the trend that participants in the critical-states condition favor the better policy, but this was not statistically significant.

We ran a one-way repeated-measures ANOVA, with policy quality as a factor and user ID as a random effect, for the question on agreement with critical states, and found a significant effect (F(1,19) = 7.92, $p = 0.01$).

\prg{Objective} We asked participants, for each of the query states (\figref{fig:drivingstates}), whether they would take control from the policy at that state. Participants in the critical-states condition consistently choose to take control more in the case of the worse policy, $\pi_B$. We do not see this trend in either of the two baseline conditions (\figref{fig:driving_obj}).

We also see that across all critical query states ($s_3$ through $s_9$), participants who saw the critical states of either policy are more likely to trust that policy and not take control of it, compared to participants who saw either a rollout of the policy or how it acts in randomly-selected states.

We ran a two-way repeated-measures ANOVA for the combination of participants' responses across all seven critical query states, and find a significant effect exposure (F(2,57) = 5.57, $p = 0.006$) and a significant interaction effect for exposure and policy quality (F(2,777) = 5.30, $p = 0.005$).
We then ran a post-hoc Tukey HSD, which showed that when participants see the critical states of $\pi_A$ and $\pi_B$, they take control significantly more for policy $\pi_B$ ($p=0.001$), but this is not true for either of the baseline conditions.

This suggests that by showing human end-users the critical states of a policy, we not only lead them to trust the policy more, but also enable them to appropriately calibrate their trust for good and not-as-good policies.
\section{Discussion and Future Work}
Our user studies suggest that showing the critical states of a policy is a promising approach for not only building trust in the policy, but also for revealing whether it is trustworthy in the first place. This can be applied to any policy trained with a maximum-entropy-based approach.

The question is, what if a policy has incorrect critical states, but it performs very well, at least in the training environment. Should we trust this policy? Or should we not trust it, because the fact that it has incorrect critical states implies that it does not truly understand the task? This is an open question for future work. Our hunch is that the latter is true---if a policy's critical states do not make sense, there are likely states (outside the training distribution) that it will not be able to generalize to. 

The primary drawback of our approach is that it places significant responsibility and mental burden on the human end-user. For instance, we assume this end-user has domain knowledge about the task; this is likely true for supervising a self-driving car or robots in a factory, but might not be true for more complex tasks. In addition, identifying false-negative critical states requires the end-user to generalize correctly about what other states the robot considers as critical, given the ones they saw. One way to address this limitation is to reason about how humans do this generalization, and show the end-user how the robot acts in additional states (critical or not) to correct their understanding.

Nonetheless, this approach of showing critical states is a step toward giving human end-users a better chance of knowing whether or not to deploy a robot, and when to take control during deployment.

\section*{Acknowledgments}
This research was funded in part by DARPA and Intel. Sandy Huang was supported by an NSF Fellowship.

\bibliographystyle{IEEEtran}
\bibliography{references}

\end{document}